\documentclass[11pt]{article}

\usepackage[margin=1.1in]{geometry}
\usepackage[T1]{fontenc}
\usepackage{lmodern}
\usepackage{amsmath,amssymb,amsthm}
\usepackage{booktabs}
\usepackage{graphicx}
\usepackage{microtype}
\usepackage{xcolor}
\usepackage[numbers,sort&compress]{natbib}
\usepackage[colorlinks=true,linkcolor=blue!60!black,citecolor=blue!60!black,urlcolor=blue!60!black]{hyperref}

\newtheorem{definition}{Definition}

\title{A Dominant Diffuse Phase in the Sparse Autoencoder \\Phase Diagram}

\author{%
  Alexis D. Plascencia\thanks{Email: \texttt{alexisd.plascencia@gmail.com}.}\\[2pt]
  \small\href{https://orcid.org/0000-0001-6393-4802}{ORCID: 0000-0001-6393-4802}%
}
\date{}

\begin{document}
\maketitle

\begin{abstract}

Sparse autoencoders (SAEs) are increasingly used to recover interpretable
features from neural-network activations, yet systematic feature
co-occurrence can cause distinct features to be \emph{absorbed} or
\emph{merged}. The MAIS-O43 open problem proposes a controlled experiment to
characterize when recovery of a true synthetic dictionary gives way to
feature merging as the nesting fraction $\gamma$, sparsity penalty
$\lambda$, and dictionary size $M$ vary. We implement the specified protocol
and evaluate 200 independently initialized fits across ten of the 165 grid
cells. We observe \textbf{zero full-dictionary recoveries and zero merges}.
Instead, every run converges to a reproducible \emph{diffuse phase}:
reconstruction is nearly perfect, but learned atoms typically remain far
from the true features (median best cosine $0.5$--$0.7$ against a $0.95$
recovery criterion) and learned codes are an order of magnitude denser than
the ground truth. This behavior persists under robustness checks and across
the full 165-cell grid using standard minibatch Adam (3{,}300 additional
fits). Since the global optimum of the exact sparse-coding objective is
known to merge nested features in the two-feature case, these results
suggest that trained SAEs need not reach the corresponding minima, and that
the phase diagram of trained models may differ fundamentally from that of
objective minimizers.

\end{abstract}

\section{Introduction}\label{sec:intro}

Sparse autoencoders have become a standard tool of mechanistic
interpretability: trained on network activations, their decoder atoms are
read as candidate \emph{features}
\citep{cunningham2023sparse,bricken2023monosemanticity}, following the
classical sparse-coding program of \citet{olshausen1996emergence} and the
superposition model of \citet{elhage2022toy}. A central failure mode is
\emph{feature absorption} \citep{chanin2024absorption}: when feature $B$
never fires without feature $A$, an SAE may learn an atom for the joint
direction rather than recovering $A$ and $B$ separately.\\

The MAIS research agenda \citep{mais2026} formalizes this phenomenon
geometrically and poses, as open problem MAIS-O43, a fully specified
experiment: measure the probabilities of \emph{recovery}, \emph{merging},
and \emph{splitting} of a known synthetic dictionary over an
$11\times5\times3$ grid of nesting fraction $\gamma$, sparsity penalty
$\lambda$, and dictionary size $M$, with 20 independent initializations per
cell (3{,}300 fits). The problem fixes the generator, the objective, the
optimizer, the sample counts, and the outcome definitions, and requires
publishing seeds and checkpoints. This paper reports the first measurements of the MAIS-O43 phase diagram, covering 10 of its 165 cells under the specified protocol: nine $M{=}256$ cells --- the full $\lambda$-sweep at $\gamma=0.5$ and the $\gamma$-row at $\lambda=10^{-2}$ for $\gamma\in{0,0.2,0.5,0.8,1.0}$, which share one cell --- plus one $M{=}512$ cell at $(\gamma{=}0.5,\lambda{=}10^{-3})$, for 200 independently initialized fits in total. These slices were chosen to resolve the two one-dimensional questions we judged most informative: the $\lambda$-response at fixed nesting, and the nesting response at the strongest penalty. The remaining 155 canonical cells are unmeasured under the full-batch protocol; all 165 cells are, however, covered by the minibatch companion grid of Section~\ref{sec:probes}.\\

%\paragraph{Contributions.}
%\begin{enumerate}
%  \item An exact, preregistered, independently testable implementation of
%        the MAIS-O43 protocol, with adversarially validated outcome
%        classifiers (Section~\ref{sec:methods}).
%  \item The first measured cells of the phase diagram
%        (Section~\ref{sec:results}): \emph{zero} recoveries and \emph{zero}
%        merges in 200 converged fits (pooled one-sided binomial 95\% upper
%        bound $\approx 1.5\%$), with a characterization of the diffuse phase
%        that occurs instead.
%  \item Two robustness probes and a complete off-protocol companion grid
%        (Section~\ref{sec:probes}) showing that the diffuse phase is
%        neither an under-training artifact (doubling the update budget
%        changes nothing) nor an artifact of the protocol's full-batch
%        optimizer (standard minibatch Adam reaches the same plateau
%        $64\times$ faster, and covers the entire 165-cell grid without a
%        single recovery or merge), and a discussion of the implication:
%        trained SAEs do not reach the global minimizers whose merging
%        behavior the theory predicts (Section~\ref{sec:discussion}).
%\end{enumerate}

This paper is organized as follows: in Section~\ref{sec:related} we situate this work in the dictionary-learning, sparse-autoencoder, and hierarchical-features literature. In Section~\ref{sec:protocol} we state the MAIS-O43 protocol: the nested-support generator, the centered SAE objective and optimizer, and the exact recovery, merge, and split definitions. In Section~\ref{sec:methods} we describe the exact, preregistered, independently testable implementation of the MAIS-O43 protocol, with adversarially validated outcome classifiers.
In Section~\ref{sec:results} we present the results: zero full-dictionary recoveries and zero merges in 200 fits, with low matched-feature fractions despite near-perfect reconstruction, and a characterization of the diffuse phase that occurs instead. \\

In Section~\ref{sec:probes} we discuss two robustness probes and a complete minibatch (off-protocol) companion grid,
demonstrating that the diffuse phase is
neither an under-training artifact (doubling the update budget
changes nothing) nor an artifact of the protocol's full-batch
optimizer (standard minibatch Adam reaches the same plateau
$64\times$ faster, and covers the entire 165-cell grid without a
single recovery or merge). In Section~\ref{sec:discussion}, we discuss the implications of trained SAEs not necessarily reaching the global minimizers of the exact sparse-coding objective whose merging behavior is predicted by theory. 
Finally, in Section~\ref{sec:conclusions} we present our conclusions.\\

\section{Related work}\label{sec:related}

\paragraph{Dictionary learning and structured sparsity.} The data model we
measure, $y=\Phi x$ with sparse nonnegative $x$, is the classical
dictionary-learning model. \citet{olshausen1996emergence} introduced it as an
account of simple-cell receptive fields; \citet{mairal2010online} gave its
modern formulation --- data as sparse linear combinations of learned
dictionary atoms --- together with scalable online algorithms and
applications across signal processing and matrix factorization. The
parent--child structure in the MAIS-O43 generator also has a classical
precedent: \citet{jenatton2011proximal} studied \emph{tree-structured}
sparse coding, in which hierarchical relationships among atoms are imposed
on the support of $x$ via structured regularization. The MAIS generator
encodes the same child-implies-parent support logic, but places it in the
\emph{data-generating process} and asks whether an unstructured learner
recovers it, rather than building it into the learner.

\paragraph{Superposition and sparse autoencoders.}
\citet{elhage2022toy} connected sparse coding to modern interpretability:
synthetic sparse features, compressed into fewer dimensions than features,
are represented by neural networks in superposition, raising the question
of whether the original features can be un-mixed. Sparse autoencoders are
the dominant answer: \citet{bricken2023monosemanticity} framed SAEs
explicitly as dictionary learning on network activations, and
\citet{cunningham2023sparse} showed the resulting features support causal
analyses of model behavior. Subsequent work has repeatedly modified the
sparsity mechanism itself: gated SAEs decouple direction selection from
magnitude estimation specifically to remove the systematic
\emph{activation shrinkage} induced by the $\ell_1$ penalty
\citep{rajamanoharan2024gated}, and top-$k$ SAEs abandon the $\ell_1$
penalty for direct sparsity control, with clean scaling laws in dictionary
size \citep{gao2024scaling}. These design responses are relevant context
for our results: MAIS-O43 fixes the \emph{vanilla} ReLU + $\ell_1$
architecture, and the collapse we measure at the largest penalty
(Section~\ref{sec:results}) is a quantitative image of the shrinkage
pathology these variants were built to avoid.

\paragraph{Hierarchical features, absorption, and hierarchical SAEs.}
The nesting axis of MAIS-O43 is motivated by \emph{feature absorption}:
\citet{chanin2024absorption} showed that when a broad feature co-occurs
with a more specific one (\emph{animal} whenever \emph{dog}), sparsity
pressure can drive an SAE to represent ``animal-except-dog'' plus ``dog''
rather than the true pair --- precisely the failure a controlled nested
generator is designed to detect. That hierarchy is the realistic case, not
a corner case, is supported by \citet{park2024geometry}, who find that
categorical and hierarchically related concepts occupy structured
geometry (simplices and orthogonal complements) in LLM representations.
The generator closest to ours appears in \citet{bussmann2025matryoshka}:
a tree of binary ground-truth features in which a child is sampled only
when its parent is active and active directions are summed --- essentially
our $A_j=B_iB_j$ --- used to show that standard SAEs absorb parent
features and that nested (Matryoshka) dictionaries mitigate this.
\citet{costa2025flat} likewise construct synthetic hierarchical generative
processes and show that a matching-pursuit-style sequential encoder
captures hierarchical and conditionally orthogonal structure that flat
SAEs miss. Most recently, \citet{zhang2026geometric} give a set-theoretic
and geometric account of concept learning in SAEs. Their framework
distinguishes graded notions of detection, separation, and approximation,
establishes capacity limits in dictionary size, and offers a unified
explanation of splitting, absorption, and hierarchical concepts. It also
predicts the effects of SAE size and sparsity that we measure empirically
here. A growing architectural line builds the hierarchy into the
learner itself: \citet{muchane2025hierarchical} add explicit semantic
hierarchy to the SAE architecture; \citet{cao2026tree} embed a tree over
the feature set and argue that \emph{activation coverage} --- the
child-active-implies-parent-active property that our generator encodes ---
is by itself insufficient to certify semantic hierarchy, adding a
reconstruction condition; and \citet{luo2026atoms} jointly learn features
and parent--child links, yielding a feature forest rather than a flat
dictionary.

\paragraph{Where this paper sits.} Two points of tension with this
literature motivate our measurements. First, the toy-model studies above
generally report that flat SAEs \emph{do} learn feature-aligned atoms in
favorable regimes and fail in specific, structured ways (absorption,
splitting) when hierarchy is present
\citep{bussmann2025matryoshka,costa2025flat,chanin2024absorption}. 
Under
the MAIS-O43 protocol we observe something different and, to our
knowledge, not previously quantified: no full-dictionary recovery and no merge under the paper's pairwise threshold definition, but a low-alignment regime in which only a small fraction of true features (at most 3.6\% per run) have a distinct atom above cosine 0.95. 
Candidate explanations for the difference include
the regime (here $m/n=4$ with $\approx 8$ simultaneously active features,
denser than many toy setups), the fixed vanilla $\ell_1$ architecture
without dead-feature resampling or auxiliary losses
\citep{bricken2023monosemanticity,gao2024scaling}, and MAIS-O43's strict
quantitative outcome definitions (cosine $\ge 0.95$ under an injective
matching; exact residual criteria for merges) in place of qualitative
absorption diagnostics. Reconciling the qualitative absorption literature
with quantitative phase measurements is an open gap that the
full MAIS-O43 grid --- and replications of it under the SAE variants above
--- would close. The distinction matters theoretically as well:
\citet{klindt2025superposition} argue on identifiability and
compressed-sensing grounds that features in superposition are recoverable
\emph{in principle} by sparse coding; our null is therefore not an
impossibility claim but a measurement of the gap between recoverability in
principle and what a trained amortized SAE attains under a fixed protocol. Second, the theoretical side of the MAIS agenda
\citep{mais2026} characterizes global \emph{minimizers}, which provably
merge nested features in the two-feature case; our measurements bear on
whether trained SAEs reach those minimizers at all
(Section~\ref{sec:discussion}).

\section{The MAIS-O43 protocol}\label{sec:protocol}

\paragraph{Generator.} Fix $n=64$ ambient dimensions and $m=256$ true
features. The dictionary $\Phi=[v_1,\dots,v_m]\in\mathbb{R}^{n\times m}$ has
i.i.d.\ $\mathcal{N}(0,I_n)$ columns normalized to unit length; one fixed draw is used
throughout. A fixed random permutation of $[m]$ pairs consecutive entries
into 128 ordered parent--child candidate pairs. For nesting fraction
$\gamma$, the first $\lfloor 128\gamma\rfloor$ pairs are declared nested.
Each sample draws independent indicators $B_i\sim\mathrm{Bernoulli}(1/32)$;
a nested pair $(i,j)$ sets activities $A_i=B_i$, $A_j=B_iB_j$ (the child
never fires without its parent; child marginal $1/1024$), all other features
keep $A_k=B_k$. Active coefficients are drawn from $\mathrm{Unif}[1,2]$, and
the observation is $y=\Phi x$. Each $\gamma$ uses $2^{18}$ fixed training
samples and $2^{16}$ fresh evaluation samples.\\

Note that the canonical $\gamma$ intervention therefore changes two things at once: the parent--child \emph{dependence} and the child's \emph{marginal frequency}. With $q=\lfloor 128\gamma\rfloor$ nested pairs, the expected ground-truth activation density is
$$d_{\mathrm{true}}(\gamma)=\frac{1}{256}\left(\frac{256-q}{32}+\frac{q}{1024}\right),$$ which decreases from $0.03125$ at $\gamma=0$ to $0.01611$ at $\gamma=1$. Effects attributed to nesting along the $\gamma$ axis are thus joint effects of dependence and rarity; disentangling them requires the matched-frequency control discussed in Section~\ref{sec:conclusions}.\\

\paragraph{Model and training.} The centered SAE has encoder
$c(y)=\mathrm{ReLU}(Wy+b)$ and decoder $\Psi\in\mathbb{R}^{n\times M}$ with
unit-norm columns and \emph{no} output bias, trained on
\begin{equation}
  G_\lambda(\Psi,W,b)
  \;=\;
  \mathbb{E}\!\left[\tfrac12\,\lVert y-\Psi\,c(y)\rVert_2^2
  \;+\;\lambda\,\lVert c(y)\rVert_1\right]
\end{equation}
by \emph{full-batch} Adam \citep{kingma2015adam} (learning rate $10^{-3}$,
$\beta=(0.9,0.999)$, $\epsilon=10^{-8}$) for $2\times10^{5}$ updates, each
update taking the exact gradient of the full $2^{18}$-sample mean. Decoder
columns are renormalized after every step. Twenty independent
initializations are trained per cell.\\

Twenty independently seeded models are trained per cell: decoder columns are initialized from independent $\mathcal{N}(0,I_{64})$ draws normalized to unit length (drawn in float64 on CPU and cast, so initialization is device- and dtype-independent), $\Psi^{\top}$. at initialization only, and $b=0$; encoder and decoder are untied thereafter. The 20 models have independent parameter seeds but share the fixed training and evaluation datasets for their $\gamma$. Formal runs use strict float32 with TF32 disabled and fixed-order chunked gradient accumulation (chunk size $16{,}384$), taking one Adam step after the exact full-data gradient is accumulated; all outcome metrics are computed in float64.\\

\paragraph{Outcome definitions.} Let $S_{ji}=\langle\psi_j,v_i\rangle$ and
call an atom \emph{live} if it activates above $10^{-8}$ on at least one
evaluation sample.\newline

\begin{definition}[Recovery, $\varepsilon=0.05$]
$\Psi$ recovers $\Phi$ if an injection $\tau:[m]\to[M]$ exists with
$S_{\tau(i),i}\ge 1-\varepsilon$ for all $i$ (decided by exact maximum
bipartite matching).\newline
\end{definition}

\begin{definition}[Merge, $(\varepsilon,\delta)=(0.05,0.1)$]
A live atom $u$ merges features $i\neq k$ if $\exists\,\alpha,\beta\ge\delta$
with $\lVert u-\alpha v_i-\beta v_k\rVert\le\varepsilon$ while
$\langle u,v_i\rangle\le 1-\delta$ and $\langle u,v_k\rangle\le 1-\delta$
(solved by exact constrained least squares over all pairs). \newline
\end{definition}

\begin{definition}[Split, $\delta=0.05$]
Feature $i$ is split if at least two live atoms satisfy
$S_{ji}\ge 1-\delta$.
\end{definition}

\section{Methods}\label{sec:methods}

\paragraph{Implementation and validation.} The generator, model, trainer,
and classifiers are implemented in PyTorch/NumPy with a 49-test suite:
data-generation invariants (child $\Rightarrow$ parent; marginals $1/32$ and
$1/1024$; bit-exact regeneration from seeds), agreement of the training
gradients with an independent NumPy implementation at $10^{-12}$ tolerance,
exact equality of chunked and single-batch full-batch gradients, and nine
adversarial classifier fixtures (greedy-defeating matching instances,
dead-atom exclusions, threshold-boundary cases at $0.95$/$0.90$/$0.05$, and
the analytic two-feature nested optimum of the MAIS agenda). All randomness
descends from a single published master seed through purpose-keyed streams.

\paragraph{Vectorized training.} The 20 fits of a cell are trained as one
batched tensor program (stacked parameters; one Adam instance, which is
exactly per-model Adam since the loss is a sum over models and Adam is
elementwise). Full training trajectories of the vectorized program match
independently trained scalar runs at $10^{-9}$ tolerance in float64 tests.
Formal runs use strict float32 with TF32 disabled on NVIDIA RTX 4080\,SUPER
GPUs (torch 2.11.0+cu128); training a 20-seed cell takes $10.2$\,h
($M=256$) or $17.5$\,h ($M=512$).

\section{Results}\label{sec:results}

\begin{table}[t]
\centering\small
\caption{Results for the ten measured canonical cells; entries are means over 20 parameter initializations per cell. No run recovered all 256 features and no run contained a live atom satisfying the pairwise merge criterion. Treating all 200 heterogeneous runs descriptively as Bernoulli trials with a common event probability gives a one-sided 95\% zero-event bound of $1.49\%$; the corresponding per-cell bound is $13.91\%$. The column ``$\geq 0.90$'' denotes the number of true features, out of 256, whose best-aligned learned atom achieves cosine similarity at least $0.90$. The density column reports the learned activation density with, in parentheses, its ratio to the $\gamma$-dependent ground-truth density $d_{\mathrm{true}}(\gamma)$ of Section~\ref{sec:protocol}; for $M=512$ the ratio compares density fractions, not active counts.}
\label{tab:cells}
\begin{tabular}{lccccc}
\toprule
Cell & matched frac. & median cos & $\ge0.90$ & density ($\times$ true) & splits \\
\midrule
$\gamma=0.0,\ \lambda=10^{-2},\ M=256$ & 0.000 & 0.580 & 0.0 & 0.479 (15.3$\times$) & 0 \\
$\gamma=0.2,\ \lambda=10^{-2},\ M=256$ & 0.000 & 0.589 & 0.0 & 0.479 (16.9$\times$) & 0 \\
$\gamma=0.5,\ \lambda=10^{-4},\ M=256$ & 0.000 & 0.472 & 2.2 & 0.312 (13.2$\times$) & 0 \\
$\gamma=0.5,\ \lambda=3{\cdot}10^{-4},\ M=256$ & 0.001 & 0.512 & 9.6 & 0.294 (12.4$\times$) & 0 \\
$\gamma=0.5,\ \lambda=10^{-3},\ M=256$ & 0.001 & 0.572 & 15.4 & 0.303 (12.8$\times$) & 0 \\
$\gamma=0.5,\ \lambda=3{\cdot}10^{-3},\ M=256$ & 0.005 & 0.727 & 23.1 & 0.450 (19.0$\times$) & 0 \\
$\gamma=0.5,\ \lambda=10^{-2},\ M=256$ & 0.000 & 0.584 & 0.0 & 0.476 (20.1$\times$) & 0 \\
$\gamma=0.8,\ \lambda=10^{-2},\ M=256$ & 0.000 & 0.538 & 0.1 & 0.467 (24.3$\times$) & 0 \\
$\gamma=1.0,\ \lambda=10^{-2},\ M=256$ & 0.000 & 0.473 & 0.2 & 0.450 (27.9$\times$) & 0 \\
$\gamma=0.5,\ \lambda=10^{-3},\ M=512$ & 0.036 & 0.712 & 34.3 & 0.234 (9.9$\times$) & 5/20 runs \\
\bottomrule
\end{tabular}
\end{table}

\paragraph{No recovery, no merging.} Table~\ref{tab:cells} summarizes the
results from all ten cells. Under the preregistered definitions, none of the
200 fits fully recovers all 256 features, and none contains a live atom
satisfying the pairwise merge criterion. Individual-feature matches do
occur: 44 runs contain at least one matched feature, for 215 matched
assignments in total (0.42\% of the $200\times256$ possible assignments).
Splitting occurred with $M=512$, where it was detected in 5 of 20 runs.

\paragraph{The diffuse phase.} Across runs in the measured cells, the mean evaluation reconstruction loss, defined as $\tfrac12\lVert y-\hat y\rVert_2^2$ per sample, ranges from $2.7\times10^{-6}$ to $5.6\times10^{-3}$. 
Mean matched-feature fractions
range from $0$ to $0.036$ per cell and mean median best cosines from $0.472$
to $0.727$, although some individual features do exceed the $0.95$ threshold
(44 of 200 runs contain at least one). Under the $10^{-8}$ activity rule,
learned codes are $9.9$--$27.9\times$ denser than the $\gamma$-dependent
ground-truth codes. Loss trajectories are late-stage plateaus rather than
exactly flat: over the final 4{,}000 updates the relative loss change has
median $6.9\times10^{-4}$ across the 200 runs and maximum $1.5\%$; Probe~A
(Section~\ref{sec:probes}) tests substantially longer training at one
selected cell. We use \emph{diffuse phase} to denote exactly this observed
combination: near-floor reconstruction error, matched-feature fractions
below $0.04$, and code density an order of magnitude above the ground truth.

\begin{figure}[t]
\centering
\includegraphics[width=0.42\textwidth]{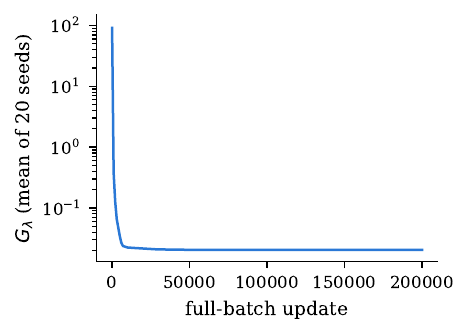}\hfill
\includegraphics[width=0.42\textwidth]{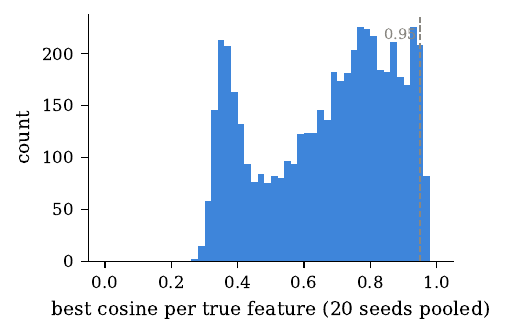}
\caption{\textit{Left panel:} Training objective for the $(\gamma{=}0.5,\lambda{=}10^{-3},M{=}512)$ cell, averaged over 20 seeds; the shaded region denotes the min--max band across seeds (which is very narrow). \textit{Right panel:} Best cosine similarity for each true feature at convergence, pooled across the cell's 20 seeds. The dashed line indicates the $0.95$ recovery criterion.}
\label{fig:convergence}
\end{figure}

\begin{figure}[t]
\centering
\includegraphics[width=0.9\textwidth]{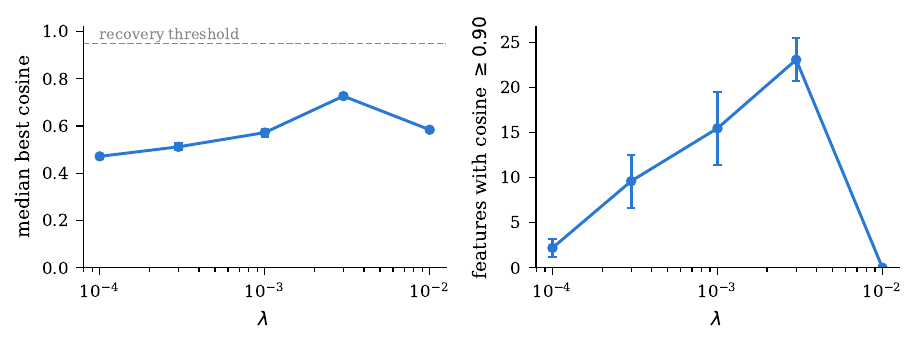}
\caption{The $\lambda$-response at $\gamma=0.5$, $M=256$ (mean $\pm$ s.d.\
over 20 seeds). \textit{Left panel:} median best cosine per feature. \textit{Right panel:}  number of
features (of 256) whose best atom reaches cosine $0.90$. The response peaks
at the interior value $\lambda=3\times10^{-3}$ and collapses at
$\lambda=10^{-2}$.}
\label{fig:lambda}
\end{figure}

\paragraph{An interior alignment maximum in $\lambda$.}
Among the five tested penalties at $\gamma=0.5$ (Figure~\ref{fig:lambda}),
the alignment summaries increase from $\lambda=10^{-4}$ to a maximum at
$\lambda=3\times10^{-3}$ (median best cosine $0.727$; on average 23 of 256
features above cosine $0.90$) and decline at $\lambda=10^{-2}$, where the
code becomes \emph{denser} (density $0.48$) while activation magnitudes
shrink --- a pattern consistent with the known $\ell_1$ activation-shrinkage
effects that gated and top-$k$ SAE variants were designed to remove
\citep{rajamanoharan2024gated,gao2024scaling}. These five measurements do
not isolate shrinkage as the cause of the alignment decline, and the best
penalty in the grid still misses the recovery criterion by a wide margin.\\

\begin{figure}[t]
\centering
\includegraphics[width=0.45\textwidth]{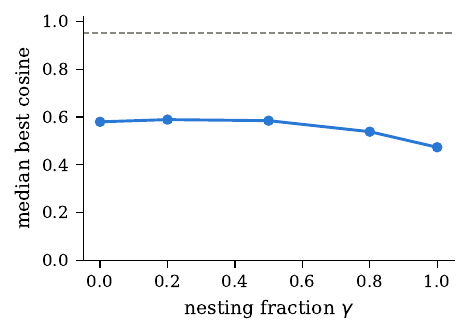}
\caption{Median best cosine as a function of the nesting fraction for $\lambda=10^{-2}$ and  $M=256$ (mean $\pm$
s.d.\ over 20 seeds).}
\label{fig:gamma}
\end{figure}

\paragraph{Nesting has almost no effect.} Along the $\lambda=10^{-2}$ row
(Figure~\ref{fig:gamma}), $\gamma\le0.5$ cells are statistically
indistinguishable, with a mild decline in alignment at $\gamma\ge0.8$.
Crucially, the merging that nesting is predicted to force never occurs
(not even at $\gamma=1$) where every feature belongs to a nested pair.

\paragraph{Dictionary size is the strongest lever.} At identical
$(\gamma,\lambda)=(0.5,10^{-3})$, doubling $M$ from 256 to 512 raises the
matched-feature fraction from $0.001$ to $0.036$ and the median cosine from
$0.572$ to $0.712$, a larger effect than any $\lambda$ or $\gamma$ change
measured. $M=512$ is also the only setting producing splits (5 of 20 runs).

\section{Robustness probes}\label{sec:probes}

Two deliberate, clearly labeled protocol deviations test the two most likely
mundane explanations of the diffuse phase; both use the peak cell
$(\gamma{=}0.5,\ \lambda{=}3\times10^{-3},\ M{=}256)$.

\begin{figure}[t]
\centering
\includegraphics[width=0.95\textwidth]{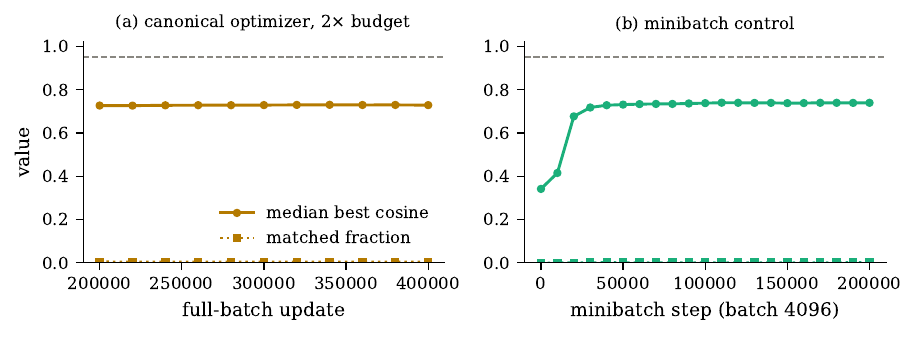}
\caption{Robustness probes on the peak cell
$(\gamma{=}0.5,\ \lambda{=}3\times10^{-3},\ M{=}256)$; mean over 20 seeds,
dashed line at the $0.95$ recovery criterion. (a)~Probe~A: continuing the
canonical full-batch runs to twice the specified update budget leaves both
the median best cosine and the matched-feature fraction flat. (b)~Probe~B:
standard minibatch Adam (off-protocol control) climbs to the same plateau
within ${\sim}60$k steps and stays there.}
\label{fig:probes}
\end{figure}

\paragraph{Probe A: is it under-training?} We continue the cell's 20 fits
from their canonical 200k-update endpoint to 400k full-batch updates, with
geometry snapshots every 20k updates (Figure~\ref{fig:probes}a). Doubling
the budget barely moves the alignment metrics: median best cosine goes from
$0.727$ to $0.729$, the number of features above cosine $0.90$ stays at
${\sim}23$, about one feature per run sits above $0.95$, and the
matched-feature fraction remains ${\sim}0.004$ throughout. Final outcomes
over the 20 continued fits: recovery $0/20$, merges $0/20$. At this cell,
then, the diffuse solution is not a snapshot of slow progress toward
recovery. This argues against simple under-training here, but does not by
itself establish stationarity or convergence at every measured cell.

\paragraph{Probe B: is it the full-batch optimizer?} We retrain the same
cell configuration --- same data, same 20 initialization seeds, same
learning rate, betas, epsilon, and decoder renormalization --- replacing
only the optimizer semantics: standard minibatch Adam with random batches of
4{,}096 samples (with replacement) for $2\times10^{5}$ steps
($\approx$3{,}100 epochs). This is how SAEs are trained in practice
\citep{bricken2023monosemanticity,cunningham2023sparse}. The result is a
strong null: minibatch training reaches the \emph{same} diffuse plateau,
only much faster. Median best cosine rises to $0.739$ by ${\sim}60$k steps
--- statistically indistinguishable from the canonical protocol's
200k-update peak of $0.727$ --- and then remains flat for the remaining
140k steps ($0.7393$ at step 200k). Final outcomes over the 20 seeds:
recovery $0/20$, merges $0/20$, matched-feature fraction $0.001$, with
${\sim}0.25$ features per run above the $0.95$ criterion. Wall-clock is
$573$\,s versus $36{,}546$\,s for the canonical cell (${\sim}64\times$
faster to the same solution). The diffuse phase therefore survives the
standard optimizer: it is a property of the objective, data regime, and
amortized encoder at these settings, not an artifact of the protocol's
full-batch semantics (Figure~\ref{fig:probes}b).

\begin{figure}[t]
\centering
\includegraphics[width=0.95\textwidth]{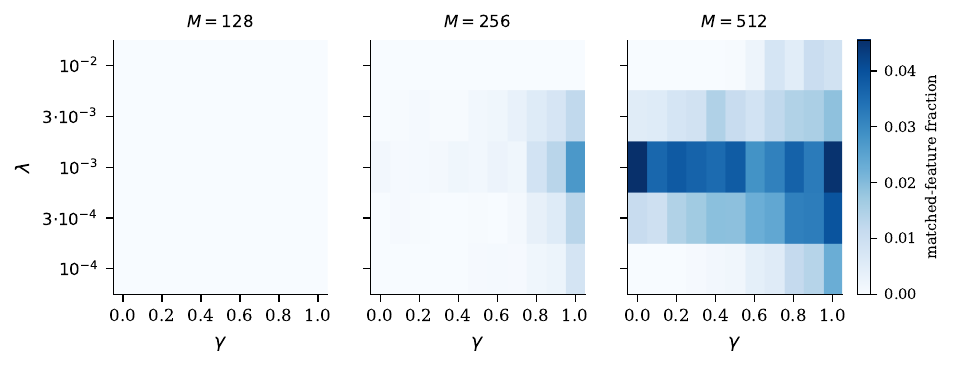}
\caption{The complete off-protocol companion grid: mean matched-feature
fraction over 20 seeds at every $(\gamma,\lambda,M)$ cell under standard
minibatch Adam (3{,}300 fits). No cell exceeds $0.046$ against a recovery
criterion requiring $1.0$; recovery and merge fractions are zero
everywhere. Alignment concentrates in the $M{=}512$, moderate-$\lambda$
corner and is essentially independent of the nesting fraction $\gamma$.}
\label{fig:mbgrid}
\end{figure}

\paragraph{The full off-protocol companion grid.} Probe~B's $64\times$
speedup makes the full grid inexpensive to measure: we retrained the
\emph{entire} 165-cell grid --- all eleven $\gamma$ values, all five
$\lambda$ values, all three dictionary sizes, 20 seeds each, 3{,}300 fits
--- under the minibatch optimizer, in $29.5$ GPU-hours of elapsed time. The
result extends the diffuse-phase finding from a slice to the whole space:
\textbf{zero recoveries and zero merges in all 3{,}300 fits}. For the
$1{,}100$ $M{=}128$ fits, full recovery is structurally impossible since
$M<m$, so the $M{=}256$ and $M{=}512$ cells are the informative recovery
tests. The only structure anywhere is the familiar $M=512$ signature ---
occasional splits (34 cells, at most $30\%$ of runs) and matched-feature
fractions peaking at $4.6\%$ at $(\gamma{=}0,\lambda{=}10^{-3},M{=}512)$
(Figure~\ref{fig:mbgrid}). The near-$\gamma$-invariance of the canonical
slice holds across the grid: the best-aligned cells at $\gamma=0$ and
$\gamma=1$ have similar peak matched fractions ($0.046$ vs $0.045$), though
this does not establish causal irrelevance of nesting or equivalence across
all alignment metrics. Being off-protocol, this companion grid does not
resolve MAIS-O43; it does, however, make the canonical grid's likely
outcome --- and the economics of confirming it --- concrete.

\section{Discussion}\label{sec:discussion}

\paragraph{A third phase.} MAIS-O43 frames its outcome space as recovery
versus merging (with splitting as a secondary effect). The sampled region of
the grid exhibits neither: a converged diffuse phase dominates, in which the
SAE solves the reconstruction problem with a dense, well-spread frame and
the sparsity penalty rotates it only slowly and incompletely toward the true
dictionary. The two-phase training dynamic is visible in every loss curve:
reconstruction collapses within a few thousand updates; the $\ell_1$ term
then decays extremely slowly as directions rotate.

\paragraph{Trained SAEs versus minimizers.} The MAIS agenda proves, for a
two-feature population under the \emph{exact-coding} objective $F_\lambda$,
that the globally optimal dictionary \emph{merges}: the optimum places one
atom on the parent and one on the normalized joint direction. Our experiment
instead trains the amortized ReLU objective $G_\lambda$ on a finite
256-feature dataset. The absence of a threshold-defined pairwise merge in
the trained models therefore does not show that optimization missed a global
minimum of $G_\lambda$; it shows that these trained many-feature amortized
solutions differ from the two-feature exact-coding benchmark. Whether that
difference is due to amortization (the concern of open problem MAIS-O39),
the many-feature distribution, finite-sample effects, or the optimization
path itself remains open. Either way, theory calibrated on exact-coding
minimizers need not transfer to trained amortized SAEs without further
argument.

\paragraph{Practical reading.} For interpretability practice the relevant
warning is not absorption but its precursor: under sufficiently unfavorable
regimes an SAE can reconstruct with very low error while only a small
minority of true features receive a distinct atom above a stringent
alignment threshold. Reconstruction quality alone therefore does not certify
dictionary recovery. This complements, rather than contradicts, the
hierarchical-SAE literature: architectures such as Matryoshka SAEs, MP-SAE,
hierarchical SAEs, Tree SAE, and feature forests
\citep{bussmann2025matryoshka,costa2025flat,muchane2025hierarchical,cao2026tree,luo2026atoms}
are motivated by flat SAEs mishandling hierarchy \emph{given} that atoms
roughly align with features; our measurements identify a regime in which
that alignment is largely absent, leaving hierarchy-aware remedies little
structure to organize. Whether those architectures raise the low matched
fractions under the MAIS-O43 generator is an open and, given Probe~B's cost
figures, inexpensive question.

\section{Conclusions}\label{sec:conclusions}

In this article we present the first measurements of the MAIS-O43 sparse-autoencoder
phase diagram. The results are unambiguous and establish four main findings.\\

First, within the sampled region of the protocol grid we find neither of the
anticipated phases.
Across 200 fits under the exact canonical protocol,
ten $(\gamma,\lambda,M)$ cells spanning the full $\lambda$ range, half the
$\gamma$ range, and two dictionary sizes, there is not a single recovery
and not a single merge under the preregistered definitions.
The corresponding pooled one-sided binomial 95\% upper bound $\approx1.5\%$ for each event.
What occurs instead is
a phase the problem statement did not anticipate: a \emph{diffuse} solution
that reconstructs the data essentially perfectly while every learned atom
remains bounded away from every true feature, with codes roughly
10--28$\times$ denser than the generating process. \\

Second, the diffuse phase is a genuine attractor of training rather than an
artifact. It is reached by all 20 independent initializations in every cell,
with final losses agreeing to three decimal places. It also survives a
doubling of the already-large canonical update budget without measurable
change (Probe~A, Section~\ref{sec:probes}) and reappears across all 165 grid
cells---a further 3{,}300 fits---when the protocol's unusual full-batch
optimizer is replaced by the minibatch Adam used in SAE practice (Probe~B and
the companion grid). None of the tested variations in compute budget or
optimizer semantics escapes this regime. Within the regime examined here, the
burden of proof therefore falls on any claim that vanilla $\ell_1$ SAE
training recovers this dictionary. \\

Third, the diffuse phase has a systematic internal structure. The alignment
summaries have an interior maximum among the five tested penalties at
$\lambda=3\times10^{-3}$ and decline at $\lambda=10^{-2}$, a pattern
consistent with $\ell_1$ activation shrinkage. In the single canonical
cross-$M$ comparison, matched-feature fraction increases substantially at
$M{=}512$ ($0.001\to0.036$ at fixed $\gamma,\lambda$), whereas the largest
range in median cosine occurs along $\lambda$. The nesting fraction
$\gamma$ --- the axis the phase diagram was designed around --- has little
effect on alignment across the measured cells, including at $\gamma=1$ where
every feature pair is nested and no merge occurs; but because the canonical
$\gamma$ intervention also lowers the nested-child frequency, dependence and
rarity remain confounded. \\

Fourth, our results expose a fundamental distinction between the behavior of theoretical minimizers and that of trained SAEs. The MAIS agenda
proves that globally optimal dictionaries merge nested features in the
two-feature case, and identifiability arguments hold that superposed sparse
features are recoverable in principle. Our measurements show trained
amortized SAEs doing neither in this regime: the phase diagram of
\emph{trained} SAEs and the phase diagram of \emph{minimizers} are
different objects, and results calibrated on the latter cannot be assumed
to describe the former. For practice, the corollary deserves emphasis:
near-perfect reconstruction is compatible with \emph{zero} feature
recovery, so reconstruction quality must not be read as evidence that an
SAE has found true features. \\

The repository releases the code, the master seed, per-cell data hashes and
results, the batched initial and final checkpoints for each measured cell,
and the scripts that regenerate every figure and table. The minibatch
companion grid is inexpensive to rerun, but it uses a different optimizer;
completing the remaining canonical full-batch cells stays computationally
costly. Matched-frequency controls, additional dictionary and data draws,
exact-coding comparisons, and replication under other SAE variants are
targeted next steps. We offer these measurements as calibration targets for
the recovery conjecture and the two-feature phase diagram of the MAIS agenda,
and as a caution for interpretability practice.\\

In summary, we studied sparse autoencoders at scale, with 256 true features
superposed in 64 dimensions. Our experiments comprise 200 canonical
full-batch fits across ten cells, together with a complete minibatch companion
grid of 3{,}300 fits. Across both experiments, no run fully recovers the
256-feature dictionary and no atom meets the merge criterion, while
matched-feature fractions remain low. This outcome lies outside the
recovery-versus-merging dichotomy around which the phase diagram was posed;
we refer to the resulting empirical pattern as a \emph{diffuse trained
regime}. The regime is reproduced across every initialization, remains
unchanged when the update budget is doubled, and reappears under minibatch
training, making it a useful calibration target. Whether it persists across
additional dictionary and data draws, SAE variants, and global minimizers of
$G_\lambda$ remains an open question. Regardless, these findings show that
theoretical predictions about optimal SAE dictionaries need not describe the
dictionaries that training in this regime actually finds.

\section{Reproducibility}\label{sec:repro}

Code, the preregistration document, per-cell results with SHA-256 data
hashes, initial and final checkpoints for all 200 fits, and the scripts that
regenerate every figure and table in this paper from those artifacts are
available at \url{https://github.com/alexisdpc/mais-o43} All randomness derives from master seed 20260805; data are
regenerated deterministically and verified against the published hashes.\\

Code, the preregistration document, per-cell result files with SHA-256 data
hashes, the batched initial and final checkpoints for the five
$(\gamma{=}0.5, M{=}256)$ $\lambda$-sweep cells, the final checkpoint of the
$M{=}512$ benchmark cell, the robustness-probe artifacts, and the scripts
that regenerate the figures and tables are available at
\url{https://github.com/alexisdpc/mais-o43}. The repository README documents
the environment and the commands used to regenerate each published figure
and table. All randomness derives from master seed 20260805; datasets are
regenerated deterministically and verified against the published hashes.\\

\subsection*{Acknowledgments}
We acknowledge the use of Claude Fable 5 in developing the numerical implementation (all outputs were verified by a human).

\bibliographystyle{unsrtnat}
\bibliography{references}

\end{document}